\documentclass{article}

\usepackage{PRIMEarxiv}
\usepackage[utf8]{inputenc} 
\usepackage[T1]{fontenc}    
\usepackage{hyperref}       
\usepackage{url}            
\usepackage{booktabs}       
\usepackage{amsfonts}       
\usepackage{nicefrac}       
\usepackage{microtype}      
\usepackage{lipsum}
\usepackage{fancyhdr}       
\usepackage{graphicx}
\usepackage[round]{natbib}
\usepackage{amsmath}

\usepackage{lscape}
\usepackage{pdfpages}
\usepackage{caption} 
\graphicspath{{media/}}     
\usepackage{caption}
\usepackage{subcaption}
\title{Financial Time Series Data Augmentation with  Generative Adversarial Networks and Extended Intertemporal Return Plots
}

\author{
  Justin Hellermann \\
  School of Business and Economics \\
  Humboldt University of Berlin \\
  Berlin\\
  \texttt{justin.hellermann@hu-berlin.de} \\
   \And
  Qinzhuan Qian \\
  IT Engineer ING \\
  ING Bank N.V. \\
  Amsterdam \\
  \texttt{phyllis.qian@mendesgans.nl} \\
    \And
  Ankit Shah \\
  Senior IT Chapter Lead ING \\
  ING Bank N.V. \\
  Amsterdam \\
  \texttt{ankit.shah@ing.com} \\
}
\begin{document}
\maketitle

\begin{abstract}
Data augmentation is a key regularization method to support the forecast and classification performance of highly parameterized models in computer vision. In the time series domain however, regularization in terms of augmentation is not equally common even though these methods have proven to mitigate effects from small sample size or non-stationarity. In this paper we apply state-of-the art image-based generative models for the task of data augmentation and introduce the extended intertemporal return plot (XIRP), a new image representation for time series. Multiple tests are conducted to assess the quality of the augmentation technique regarding its ability to synthesize time series effectively and improve forecast results on a subset of the M4 competition. We further investigate the relationship between data set characteristics and sampling results via Shapley values for feature attribution on the performance metrics and the optimal ratio of augmented data. Over all data sets, our approach proves to be effective in reducing the return forecast error by 7\% on 79\% of the financial data sets with varying statistical properties and frequencies.
\end{abstract}

\keywords{Data Augmentation\and Generative Adversarial Networks \and Time Series \and Extended Intertemporal Return Plots \and XIRP  \and Intertemporal Return Plots \and IRP \and Shapley Values}

\section{Introduction}
Augmenting image data has become a fundamental part of many machine learning pipelines and is a widely applied regularization technique to improve model robustness and performance  \citep[p.~309]{geron_hands_machine_2017}. In the computer vision community these techniques magnify the number of training examples by scaling, rotating and shifting images. This allows for running multiple training epochs on variations of the same underlying image.\newline
Even though time series also face the challenge of limited training examples, the application of data augmentation methods is not equally common. In the classic, parametric setting, data augmentation is not per se required, since the models have few parameters to fit. Nevertheless, this changes quickly, once more complex neural network based models have to be trained. But data augmentation has more far-reaching benefits than solely allowing to train highly parameterized models. Expanding the amount of train data can improve the performance of regression and classification tasks, analogously to the improved estimates via bootstrapping techniques. In physics and system engineering the benefit of time series augmentation is established and applied successfully to enlarge existing train data \citep{theiler1992testing}. This partly results from the characteristics of time series in the context of system engineering which often exhibit less randomness compared to economic and financial data sets.\newline
For decades, data augmentation therefore remained in the area of physics before moving into finance \citep{le2016data,ziyin2021data}. Just recently, researchers began to apply data augmentation methods to financial time series trying to mitigate the effects of random and noisy features, sudden jumps and nonstationary. Since many nonparametric time series models need increased amounts of data to generalize well, augmenting existing time series is promising to boost performance, as shown by \cite{yuan2020using,lee2020stock}. In this paper we consider different financial data sets and introduce an extended version of the IRP to create image representations encoding both the time series as well as intertemporal returns. Combining the new image representation with advanced generative models facilitates the augmentation of time series data. The new approach is tested on a large number of financial datasets with varying frequencies spanning from daily to yearly data. A series of test is conducted in and among the different data frequencies to answer the following research questions:
i) Can forecasting models benefit from augmented data and if so to what extent?
ii) What is the optimal ratio between real and augmented data? 
iii) Which statistical properties contribute to a successful augmentation?

\section{Related Work}
Since the proposed approach combines generative models, image representations of time series and data augmentation, we discuss related work at different topical intersections. The related work mainly focuses on time series data augmentation methods and its the benefits as well as an introduction to other time series image representations.\newline
Data augmentation for the purpose of classification has been applied by  \cite{forestier2017generating}, who use dynamic time warping barycentric averaging to produce new samples. A more complex approach is chosen by \cite{le2016data}, who use a deep convolutional neural network architecture to generate synthetic data for time series classification. For time series forecasting many models rely on transformation and bootstrapping methods \cite{iftikhar2017scalable,bergmeir2016bagging,BANDARA2021108148}. \cite{bergmeir2016bagging} use repeated sampling techniques to improve forecasts by aggregation (bagging) of exponential smoothing methods. By separating the time series into the trend, seasonal part, and remainder bagging using Box–Cox transformation and STL decomposition, their approach returned promising results for monthly economic data. The related concept of IAAFT by \cite{schreiber2000346} combines the bootstrap approach with shuffling time series  which have been generated by Fourier transform to simulate new time series. From the area of parametric simulations \cite{denaxas2015syntise,papaefthymiou2008mcmc} apply Markov Chain Monte Carlo methods to simulate and thereby augment time series data. Another recent approach named GRATIS by \cite{kang2020gratis}, use mixture autoregressive models to generate time series with diverse and controllable features. According to the authors, the main benefit lies in a more reliable model selection due to the possibility to test on an extended amount of data. Apart from mixture autoregressive models which require assumptions on the underlying process, the non-parametric class of GAN architectures, especially the one proposed by \cite{karras2020analyzing}, have received significant attention, as a result of their outstanding successes in generating new images. The successes of the adversarial methods were also partly transferred to the domain of time series to synthetic data \citep{fu2019time,NEURIPS2019_c9efe5f2,esteban2017realvalued,zhang2015character}\newline
Studies regarding the augmentation benefit for forecasting are sparsely covered. \cite{BANDARA2021108148} evaluated the GRATIS and Moving Block Bootstrapping, Dynamic Time Warping Barycentric Averaging techniques to improve the accuracy of global forecasting models. \cite{smyl2016data} examine multiple time series augmentation methods (Markov Chain Monte Carlo, local-global-trend algorithm, exponential smoothing) for multiple short time series forecasting with recurrent neural networks.\newline
Related image representations are Gramian Angular Summation and Gramian Angular Difference Fields \citep{Wang2014EncodingTS}, which rely on a polar encodings their inverse trigonometric function to the sum or the difference of the encoding. Also Waveforms are often used to encode the shape of a time series graph as a function of time. Gramian Angular Fields, Waveforms and recurrence plots rely on a deterministic encodings of a time series. A different approach is chosen by Markov transition fields as applied by \cite{Wang2014EncodingTS} to bin the time series in different classes and then calculate transition probabilities and represent them in a two dimensional space. Also spectogramms rely on a probablistic encoding and have been applied successfully in processing audio signals and visualizing the spectrum of frequencies of a signal as it varies with time \citep{wyse_audio_2017} while each plot is characteristic for a certain sound or a spoken word. 
Depending on the approach of encoding, some of the upper methods are limited in their applications and can either recover the initial time series only partially or approximately.\newline
This paper complements the existing studies by two novel aspects. First, the paper introduces a new image representation named XIRP and then trains an image-based GAN to augment the existing data. Further, we apply Shapley values for model explainability to get an improved insight into what factors contribute to each of the performance metrics. The ensemble of Shapley values and image-based generative models for time series is, to the best of our knowledge, not yet covered in the literature.
\section{Methodology}
In this paper we use a Wasserstein GAN which is an extension of the vanilla GAN by \cite{goodfellow_generative_2014}. The vanilla GAN consists a generator network $G$ and a discriminator network $D$ with opposing objective functions. In the course of training, $G$ generates a fake distribution $\tilde{x}$ and proposes it to $D$, which assesses the likelihood of $\tilde{x}$ originating from the true data distribution. The adversarial game between $G$ and $D$ can be described by the objective function
\begin{equation}
\min_{G} \quad \max_{D}\quad \mathop{\mathbb{E}}_{x\sim \mathop{\mathbb{P}_r}}[log(D(x))]+\mathop{\mathbb{E}}_{\tilde{x}\sim \mathop{\mathbb{P}_g}}[log(1-D(\tilde{x}))]
\end{equation}
where $D$ and $G$ try to maximize and minimize (3), respectively. $D(x)$ is the estimate of the discriminator that $x$ is real while $\mathbb{P}_r$ denotes the real data distribution. The generated data is denoted by $\tilde{x}$ with $\tilde{x}=G(z)$ and $z\sim p(z)$ where $p$ denotes some distribution, such as uniform or Gaussian. $\mathbb{P}_g$ is the distribution of the generated data, or, put differently, the model distribution.
Having the opposing objectives, both networks play an iterative game and train each other in the process. 
The above vanilla GAN suffers from stability problems and is difficult to train \citep{goodfellow_generative_2014}. \cite{arjovsky_wasserstein_2017} address this problem by proposing a Wasserstein-GAN (WGAN) with a more stable training behavior including correlation between the value of the objective function and sample quality. An intuitive explanation for the Wasserstein distance as a loss function can be obtained from its synonym \textit{Earth Mover Distance}. It can be related to the minimum cost of transporting mass in order to transform the distribution $q$ into the distribution $p$, whereby the cost is mass times transport distance. The Wasserstein metric changes the upper objective function to
\begin{align}
\min_{G} \quad \max_{D\epsilon \mathcal{D}}\quad \mathop{\mathbb{E}}_{x\sim \mathop{\mathbb{P}_r}}[D(x)]-\mathop{\mathbb{E}}_{\tilde{x}\sim \mathop{\mathbb{P}_g}}[D(\tilde{x})]
\end{align}
where $\mathcal{D}$ is defined as a set of 1-Lipschitz functions and $\hat{x}$ is a randomly sampled subset. Instead of classifying generated images as real or fake, the WGAN replaces the discriminator model by a critic that scores the realness or fakeness of an image. \cite{gulrajani_improved_2017} further improve training in their WGAN-GP, which incorporates gradient penalty terms for the discriminative network. This changes the objective function to
\begin{align}
\min_{G} \quad \max_{D\epsilon \mathcal{D}}\quad \mathop{\mathbb{E}}_{x\sim \mathop{\mathbb{P}_r}}[D(x)]-\mathop{\mathbb{E}}_{\tilde{x}\sim \mathop{\mathbb{P}_g}}[D(\tilde{x})]+
\lambda \mathop{\mathbb{E}}_{\tilde{x}\sim \mathop{\mathbb{P}}_{\hat{x}}} [({\| \nabla_{\hat{x}}D(\hat{x})\|}_2-1)^2]
\end{align}
The gradient penalty term $[({\| \nabla_{\hat{x}}D(\hat{x})\|}_2-1)^2]$ opposes vanishing and exploding gradients. We refer to \cite{gulrajani_improved_2017} for a more detailed explanation of all underlying concepts and considerations.
In this paper, $G$ is trained on an extended version of IRPs recently proposed by \cite{hellermann2021leveraging}. The concept of IRPs mixes insights from quantitative finance with the idea recurrence plots by \cite{eckmann_recurrence_1987}. It calculates returns between all points in a time series returning an $NxN$ representation of a time series with length $N$. The IRP with indexes $i$ and $j$ is obtained by calculating the pairwise logarithmic return
\begin{align}
R^{C}_{i,j} =
log \left(\frac{x_i}{x_j}\right) \quad \forall i,j\in \{1,2,\ldots ,S\}
\end{align} 
Once the start value at $t=0$ is known, any time series can be recovered from the IRP regardless whether it is a real or simulated representation.  Note, that $x \in \mathbb{R}_{+}$ must be ensured by scaling since returns must be calculated on positive values.\newline
Since we want to compare augmentation techniques that do not sample returns but a scaled time series an extension is required. A careful look at the IRP reveals that in case $i=j \implies R^{C}_{i,j} = 0$ which leads to a diagonal filled with zeros. This motivates the formulation to the extended IRP (XIRP) which complements the intertemporal return structure by adding extra information to the encoding. Thus, equation (4) can be written as:
\begin{equation}
R^{X}_{i,j} =
\begin{cases}
log \left(\frac{x_i}{x_j}\right) \quad & \forall i,j\in \{1,2,\ldots ,S\} \wedge i \neq j \\
x_{i} \quad & \forall i,j\in \{1,2,\ldots ,S\} \wedge i=j
\end{cases}
\end{equation} 
The advantages of the approach are manifold. First, the diagonal of the IRP does not contain any information and is filled with additional feature in the XIRP. Second, we can directly compare the results to other sampling techniques without rescaling and calculating returns for the object of comparison. Since the diagonal of the XIRP contains an additional feature, a joint scaling would suggest a single feature and affect results negatively. As a result, the diagonal and the off-diagonal have to be scaled independently. Linking the concepts of GAN and XIRP we can now sample and extract the time series from the diagonal while the returns are saved in the off-diagonal. There are multiple ways of recovering a time series from an XIRP. Since the initial values are stored in the diagonal, these can be directly extracted, ignoring all off-diagonal information encoded in the data. However, this would neglect parts of the encoded information thus we propose to extract the diagonal in a first step and then use the diagonal as a baseline time series to calculate variants. This involves replacing the diagonal of the initial XIRP with $d^{XIRP}_{ij}=1 \quad \forall i,j\in \{1,2,\ldots ,S\}$. Afterwards, the extracted diagonal can be multiplied column-wise with the $XIRP_{d=1}$, so that $S$ time series are obtained which consider the dynamics encoded in the off-diagonal. An additional option is to average over the $S$ time series or randomly pick one of them. Randomly choosing a single instance or averaging over all time series are valid methods to obtain a single time series. Thus, the approach can be used as a scale-invariant encoding and also as an encoding for the raw time series.\newline
In order to identify the contributing factors for a successful augmentation we make use of the concept of Shapley Values. Invented as a solution mechanism for cooperative behavior in game theory by \cite{shapley_notes_1951}, it has become popular for explaining the contribution of features to a prediction and is now frequently used to enhance model understanding \citep{ghorbani2019data,merrick2020explanation,antipov2020interpretable}.
To quantify the feature importance, these are successively added in a random order and the contribution to the prediction is calculated. The Shapley value of a feature value is the average change in the prediction that results from the new feature having joint. Its ability to explain the predictions of nonlinear models along with its desirable properties make it a competitive alternative to more complex models such as Layer-Wise Relevance Propagation or Locally Interpretable Model-Agnostic Explanations (LIME). Please refer to \cite{shapley_notes_1951} for a detailed introduction to the concept, its variants and considerations.
\newpage
\section{Data and Test Design}
For the upcoming series of tests we use financial data with varying frequencies from the M4 forecasting competition of \cite{makridakis2018m4}. In order to reduce the computation volume related to the extensive backtests, we use $N=100$ data sets of each $n=20$ daily, weekly, monthly, quarterly and yearly data wile the number of elements is limited to $n_d=1000$ daily, $n_w=500$ weekly, $n_m=250$ monthly, $n_q=100$ quarterly and $n_y=75$ yearly data points. For a complete list of variables see Table 1 in the Appendix.\newline
To test the quality of the synthetic data generated by the different augmentation methods, we conduct a series of experiments. The first three experiments test for the similarity of synthetic and real instances via different approaches. The first test uses a dimensionality reduction technique to visualize synthetic and real data sequences. The second test trains a recurrent neural network on synthetic data to predict real data while the third test trains a classifier to distinguish between real and synthetic sequences. A final fourth test evaluates the benefit of the augmentation methods by training a model on varying levels of generated data.\newline
The first test applies the the Uniform Manifold Approximation and Projection (UMAP) algorithm \citep{mcinnes_umap_2020} for clustering. A similar approach has been applied by \cite{NEURIPS2019_c9efe5f2,ali_timecluster_2019} and treats each of the $d$ time steps as an individual feature. The UMAP algorithm receives sequences of length $D_{seq}=28$ and reduces the dimension to $D_{umap}=2$. Elements located closely in the 2D space are likely to have similar characteristics. Then, we plot the representation in $d=2$ and randomly choose up to $n=500$ synthetic and real sequences, depending on the size of the initial data set. The individual points are colored to indicate which are original or synthetic. In case, real and synthetic time series are clearly different, we observe two distinct clusters with small distance inside each cluster and a large difference to the other cluster(s). Else, we observe one or more mixed clusters with heterogeneous distribution of real and synthetic elements inside each cluster.
\begin{figure}[htbp]
\centering
  \begin{subfigure}[b]{0.48\textwidth}
    \includegraphics[width=\textwidth]{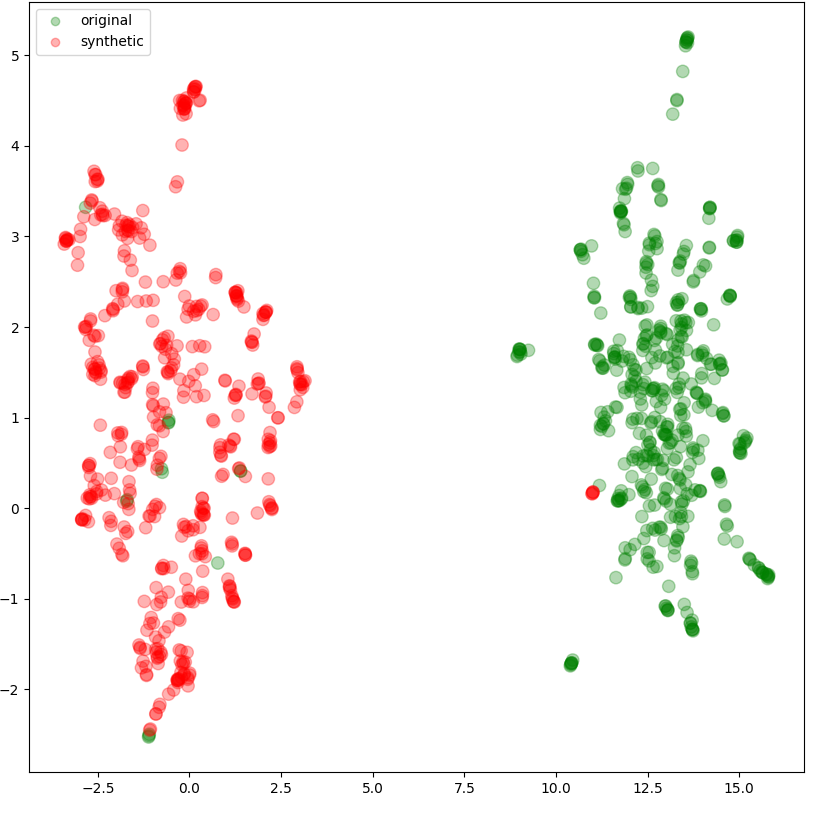}
    \caption{UMAP plot Weekly Data (W77)}
  \end{subfigure}
  \begin{subfigure}[b]{0.48\textwidth}
    \includegraphics[width=\textwidth]{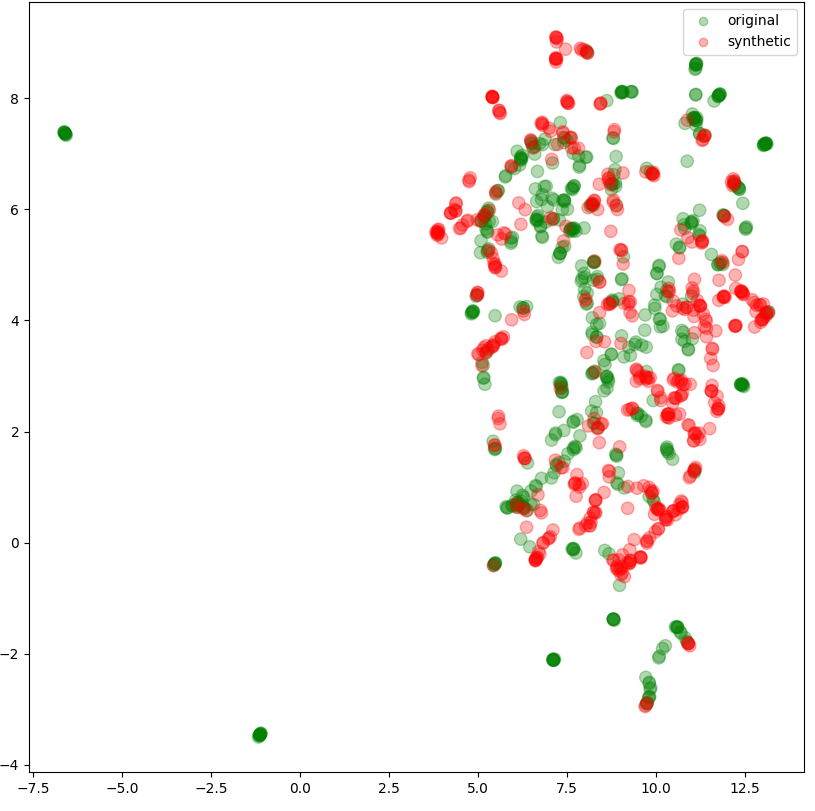}
    \caption{UMAP plot for Weekly Data (W85)}
  \end{subfigure}
    \label{umap_plots}
    \caption{Selected UMAP Results}
\end{figure}
Figure 1 (a) contains the clustering result of a weekly data set (W77). The original data (green) and the generated data (red) separate with few exceptions in two distinguishable clusters indicating deficient sample quality. Here, the samples are significantly different such that the UMAP algorithm can easily differentiate between them. An increased homogeneity in the samples is displayed Figure 1 (b). It shows real and synthetic instances of weekly data (W85), but displays a single large cluster mixed with synthetic and real instances. Except for a few outliers, synthetic and real elements mix well, while the synthetic instances have a slightly more wide-spread cluster. This is an example of a successful augmentation where the UMAP algorithm cannot distinguish between original and synthetic elements. Due to the high number of data sets we only show two exemplary UMAP plots at this point.\newline
After a first qualitative assessment, the next test performs a predictive assessment on the synthetic data. In order to get more detailed information whether the augmentation is useful for predicting the real data, we train a recurrent neural network (RNN) solely on synthetic instances before predicting the real instances. We choose a small LSTM network of $n_{hl}=3$ hidden layers with $n_{ne}=7$ neurons each. In case the augmentation method works well, a model trained on synthetic data predicts real data accurately. To assess the quality we perform a 80/20 split for train and test set. The training process is repeated $k=10$ times and uses early stopping after $c=5$ periods on the mean absolute error of the test set to counter imperfect initialization and overfitting. The predictive score $s_p$ for all data sets can be obtained from Table 1 in the Appendix section.\newline
The third, discriminative test, acts a quantitative expansion to the first qualitative test and assesses whether a generated time series is reasonably similar to the original time series. Again, a RNN is trained on sequences of length $D=28$ but now with the goal to distinguish real and synthetic instances with the same split for train and test data as well as early stopping on the binary crossentropy error. Again, we choose a small LSTM network with $n_{hl}=3$ hidden layers with $n_{ne}=8$ neurons each. The discriminative score $s_d$ can be obtained from Table 1 in the Appendix section.\newline
A fourth experiment examines the augmentation benefit for the WGAN-GP trained on the XIRPs for each of the data sets. Therefore an RNN is trained stepwise with an increasing fraction of synthetic data ranging from 0\% to 50\%. At each step a train/test split shuffles the sequences and randomly assigns them to one of the sets. The RNN is evaluated analogously to the predictive test using the same early stopping mechanism. Note, that the test and validation set always consist of real data instances to keep synthetic elements in the training set and thus avoid corrupting reliability of the performance measure. A train/test split of 80/20 is kept throughout the test. For each fraction of synthetic data $\alpha$ we repeat the test $k=10$ times to improve the stability of the results. As a reference, we choose $RMSE_{\alpha=0}$ and then use the percentage improvement compared to the optimal level $RMSE_{\alpha^{\star}}$. 
\begin{figure}
    \centering
    \includegraphics[scale=.5]{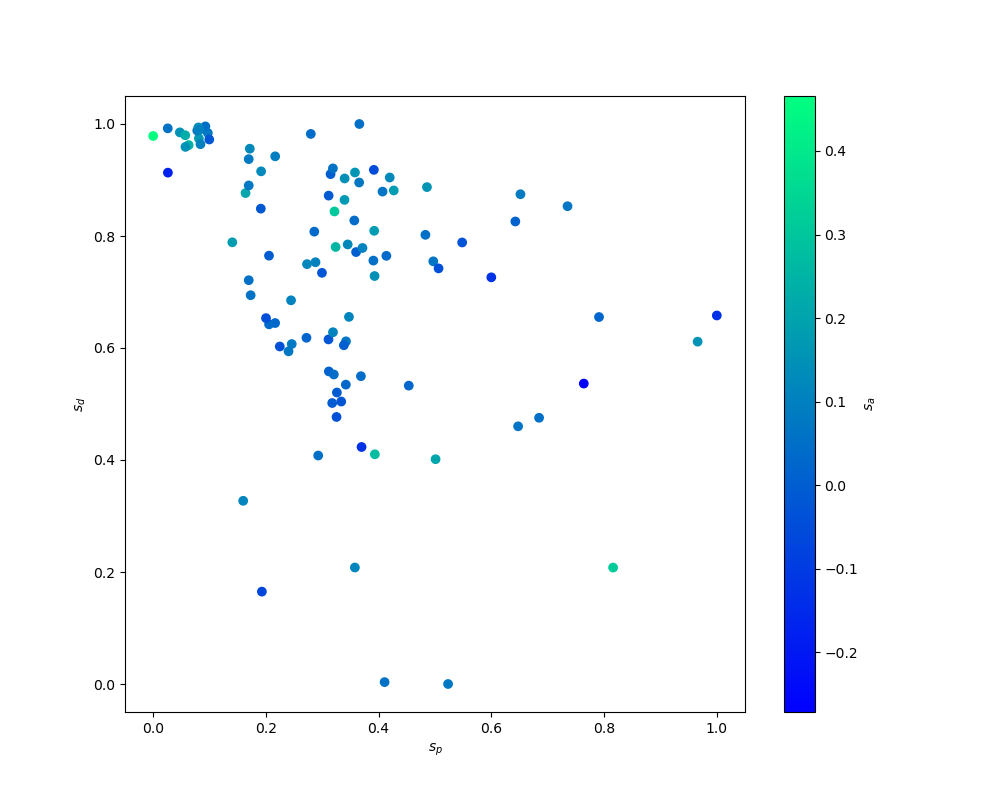}
    \caption{Relation between Predictive, Discriminative and Augmentation Score}
    \label{fig:disc_pred_aug}
\end{figure}
Figure \ref{fig:disc_pred_aug} plots the three scores $s_p$, $s_d$ and $s_a$ and their relationship.\newline
In case of perfect alignment, we could observe a clear relationship, be it linear, quadratic or kubic and see a gradular color change along either axis. A look at the relationships between the scores via the Spearman rank correlation coefficient $\rho_r$ displayed in Table 2 supports the impression of the plot, that there is a negative relationship between $s_d$ and $s_p$, namely $\rho_r(s_d,s_p)=-0.4469$.
\begin{table}
\centering
\begin{tabular}{c|c|c|c}
    $\rho $ & $s_a$ & $s_p$ & $s_d$ \\
    \hline
    & & & \\
    $s_a$ &  1   & -0.1359  &  0.2837 \\
    $s_p$ &   0.1359  &   1    &  0.4469 \\
    $s_d$ &   -0.2837  &  -0.4469     & 1\\
\end{tabular}
\caption{Spearman rank correlation between the test scores}
\end{table}
The negative sign originates from the construction of the scores, since a small predictive, high discriminative and high augmentation scores are optimal.
The relationship of the augmentation benefit towards predictive and discriminative score is less distinct. Its interaction with the predictive score $\rho_r(s_a,s_p)=-0.1359$ is a slightly negative one, while $\rho_r(s_a,s_d)=0.2837$ is slightly positive.
Since the measures account for different aspects of the generative model, they are designed to be slightly correlated but not perfectly aligned, which would indicate redundancy.\\
An important aspect when augmenting data, is to what extend we can reduce the forecast error. In a first approach, inspect the augmentation score in terms of percentage decrease in RMSE on the frequency domain. The scores aggregated over all frequencies and plotted in Figure 3 (a) shows that some data sets benefit from augmentation resulting in a decrease the error by 30\% while for other data sets, augmentation is not successful resulting in larger forecasting errors and thus a negative $s_a$. On average, the RMSE decreases by $\mu=0.07$ or 7\%. Taking a closer look at the daily data in Figure 3(b) shows that for a daily frequency the benefit diverges to either small positive/negative benefits or large benefits. For weekly data $\mu$ stays roughly unchanged and the data almost resembles a uniform distribution from $s_a=-0.05$ to $s_a=0.15$. For monthly data $\mu=0.04$ is smaller but differs little regarding the variance. Regarding the quarterly and yearly data in Figures 3(e) and 3(f) we see a slight shift in the mean towards a higher augmentation benefit. \newline
In all six frequencies we see only little differences in terms of $\mu$, which is surprising, since the observations and their data generating process are different in nature. Nevertheless, the WGAN-GP approach with XIRPs is able to improve 85\% of the daily data, 75\% of the weekly, 70\% of the monthly, 90\% of the quarterly data and 75\% of the yearly data sets.
\begin{figure}[htbp]
\centering
  \begin{subfigure}[b]{0.31\textwidth}
    \includegraphics[width=\textwidth]{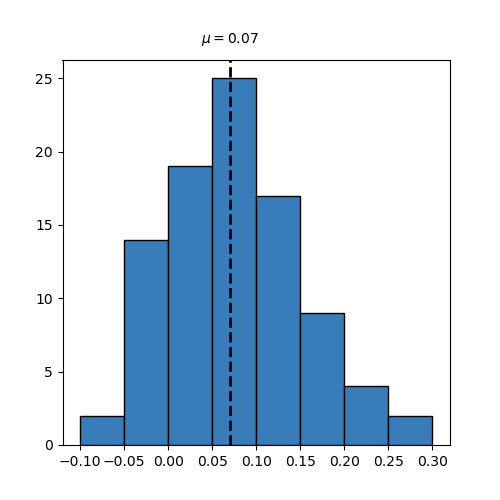}
    \caption{$s_a$ for all data sets}
  \end{subfigure}
    \begin{subfigure}[b]{0.31\textwidth}
    \includegraphics[width=\textwidth]{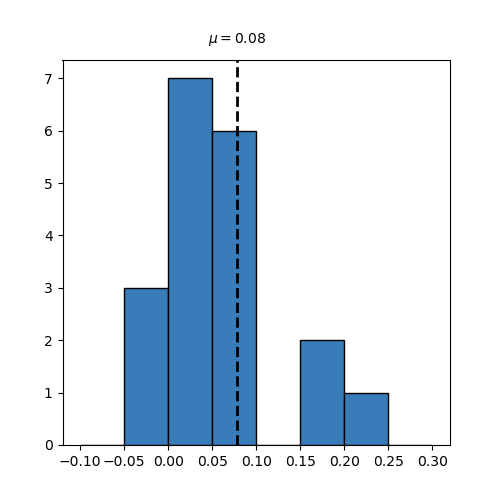}
    \caption{$s_a$ for daily data}
  \end{subfigure}
    \begin{subfigure}[b]{0.31\textwidth}
    \includegraphics[width=\textwidth]{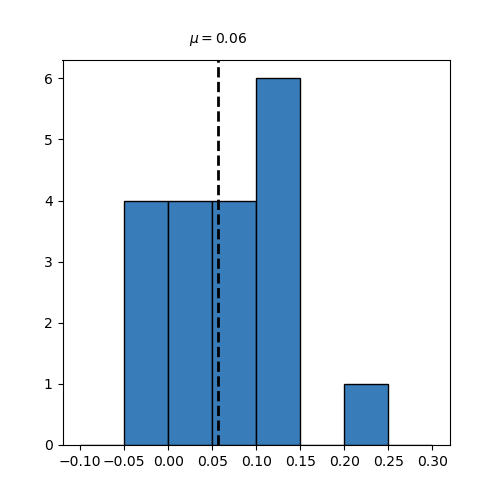}
    \caption{$s_a$ for weekly data}
  \end{subfigure}
    \begin{subfigure}[b]{0.31\textwidth}
    \includegraphics[width=\textwidth]{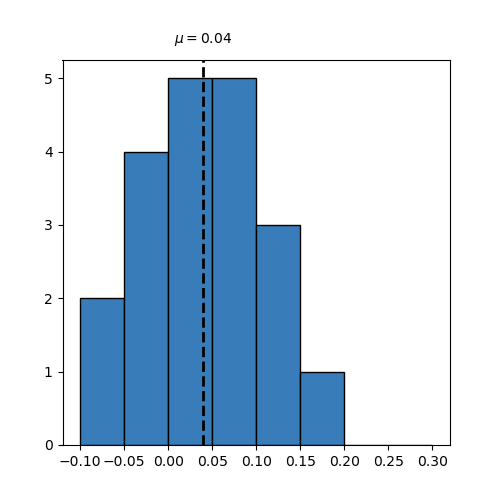}
    \caption{$s_a$ for monthly data}
  \end{subfigure}
    \begin{subfigure}[b]{0.31\textwidth}
    \includegraphics[width=\textwidth]{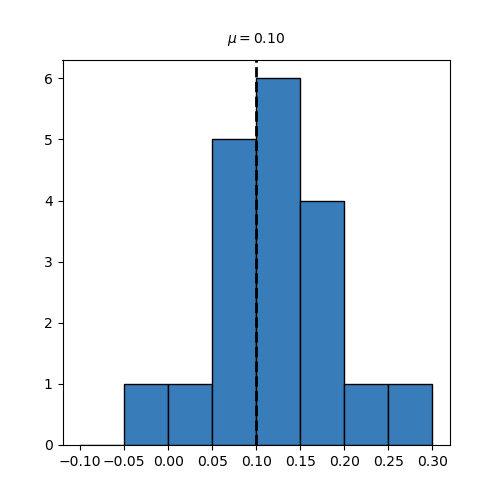}
    \caption{$s_a$ for quarterly data}
  \end{subfigure}
    \begin{subfigure}[b]{0.31\textwidth}
    \includegraphics[width=\textwidth]{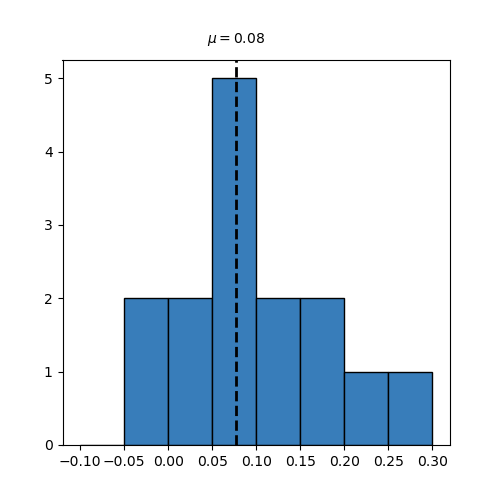}
    \caption{$s_a$ for yearly data}
  \end{subfigure}
    \label{fig:augmentation_score}
    \caption{Augmentation Score Distribution}
\end{figure}

\begin{figure}[htbp]
\centering
  \begin{subfigure}[b]{0.32\textwidth}
    \includegraphics[width=\textwidth]{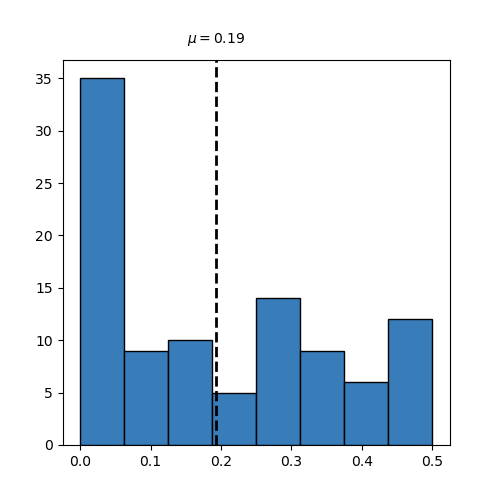}
    \caption{$\alpha$ for all data sets}
  \end{subfigure}
    \begin{subfigure}[b]{0.32\textwidth}
    \includegraphics[width=\textwidth]{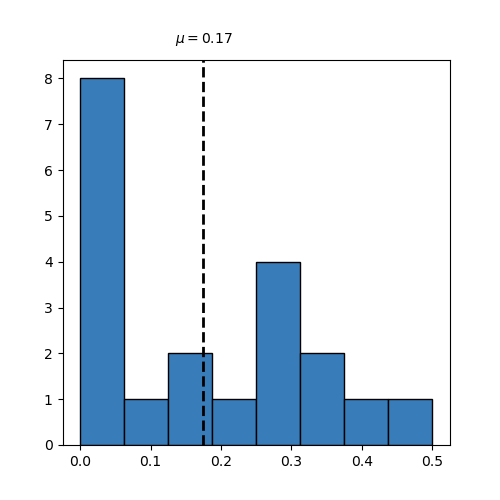}
    \caption{$\alpha$ for daily data}
  \end{subfigure}
    \begin{subfigure}[b]{0.32\textwidth}
    \includegraphics[width=\textwidth]{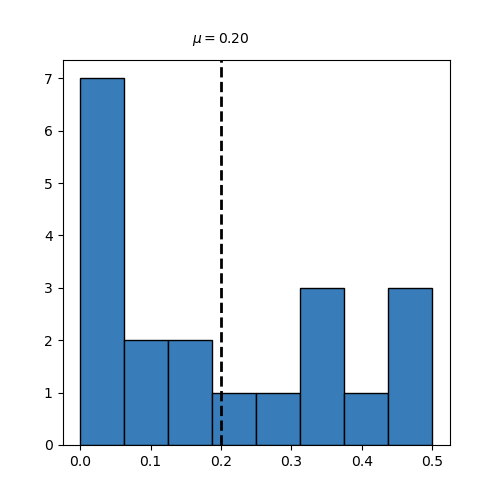}
    \caption{$\alpha$ for weekly data}
  \end{subfigure}
  
    \begin{subfigure}[b]{0.32\textwidth}
    \includegraphics[width=\textwidth]{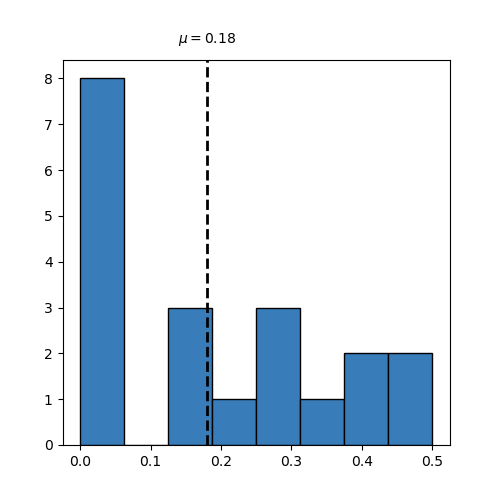}
    \caption{$\alpha$ for monthly data}
  \end{subfigure}
    \begin{subfigure}[b]{0.32\textwidth}
    \includegraphics[width=\textwidth]{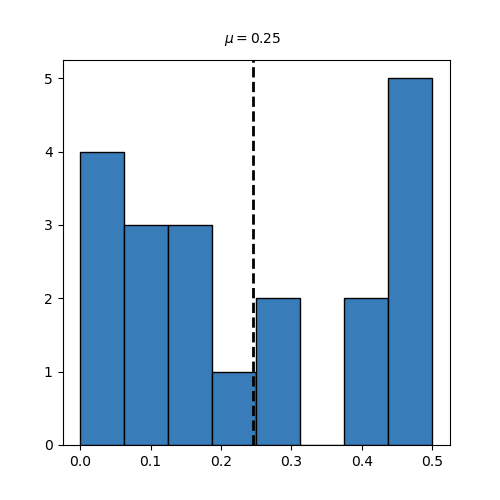}
    \caption{$\alpha$ for quarterly data}
  \end{subfigure}
    \begin{subfigure}[b]{0.32\textwidth}
    \includegraphics[width=\textwidth]{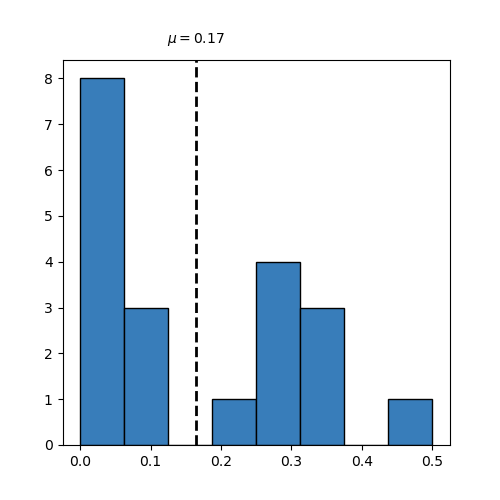}
    \caption{$s_a$ for yearly data}
  \end{subfigure}
    \label{fig:augmentation_level}
    \caption{Augmentation Level Distribution}
\end{figure}

In the following step, we explore the optimal amount of augmented data $\alpha_{\star}$ and therefore consider the distribution of $\alpha$. The results are displayed in Figure 4 where we see relatively huge bars for $\alpha=0$, which corresponds to an unsuccessful augmentation. In this case the best is to omit generated data. The results look similar for all frequencies with varying $\mu$ except of the quarterly data. The $\alpha_{\star}$ for quarterly data exhibits a large proportion of high augmentation levels, where 5 out of 20 data sets can be augmented up to a rate of 50\% of generated data. This is interesting since compared to the next lower frequency of yearly data, this is only the case for 1 in 20 data sets. Over all financial data sets, we can see that on average $\alpha_{\star}=.2$ or a level of roughly 20\% of generated data lowers the forecast error in the most effective way. \\
In Figures 3 and 4 we can see that $s_{\alpha}$ and $\alpha_{\star}$ vary in and across the different frequencies. In order to know, which factors drive not only $s_a$ but also $s_p$ and $s_d$, we apply the concept of Shapley values to a set of moments (mean $\mu$, variance. $\sigma^2$, skewness $\mu_3$ and kurtosis $\mu_4$) contained in Table 1 as well as the results from the previous discriminative and predictive tests.
Besides the descriptive statistics, we add the value of Ljung-Box test statistic to quantify whether a group of autocorrelations differ from zero \cite{ljung,box}. With $H_0$ the test assumes the data is independently distributed and thus any correlations must be of random nature. The number of lags being tested over is denoted by $h$ and we can reject $H_0$ if the test statistic\footnote{The test statistic is calculated via $Q=n(n+2)\sum _{k=1}^{h}{\frac {{\hat {\rho }}_{k}^{2}}{n-k}}$ with sample size $n$ and sample autocorrelation $\hat {\rho}$ at lag $k$} $Q > \chi_{1-\alpha,h}^2$. To incorporate the commonly present behavior of volatility clustering, the test is conducted once on the raw returns $Q_r$ and on the absolute returns $Q_{\lvert r \rvert}$ to capture large sequential movements with both signs. \newline 
\begin{figure}[htbp]
\centering
  \begin{subfigure}[b]{0.49\textwidth}
    \includegraphics[width=\textwidth]{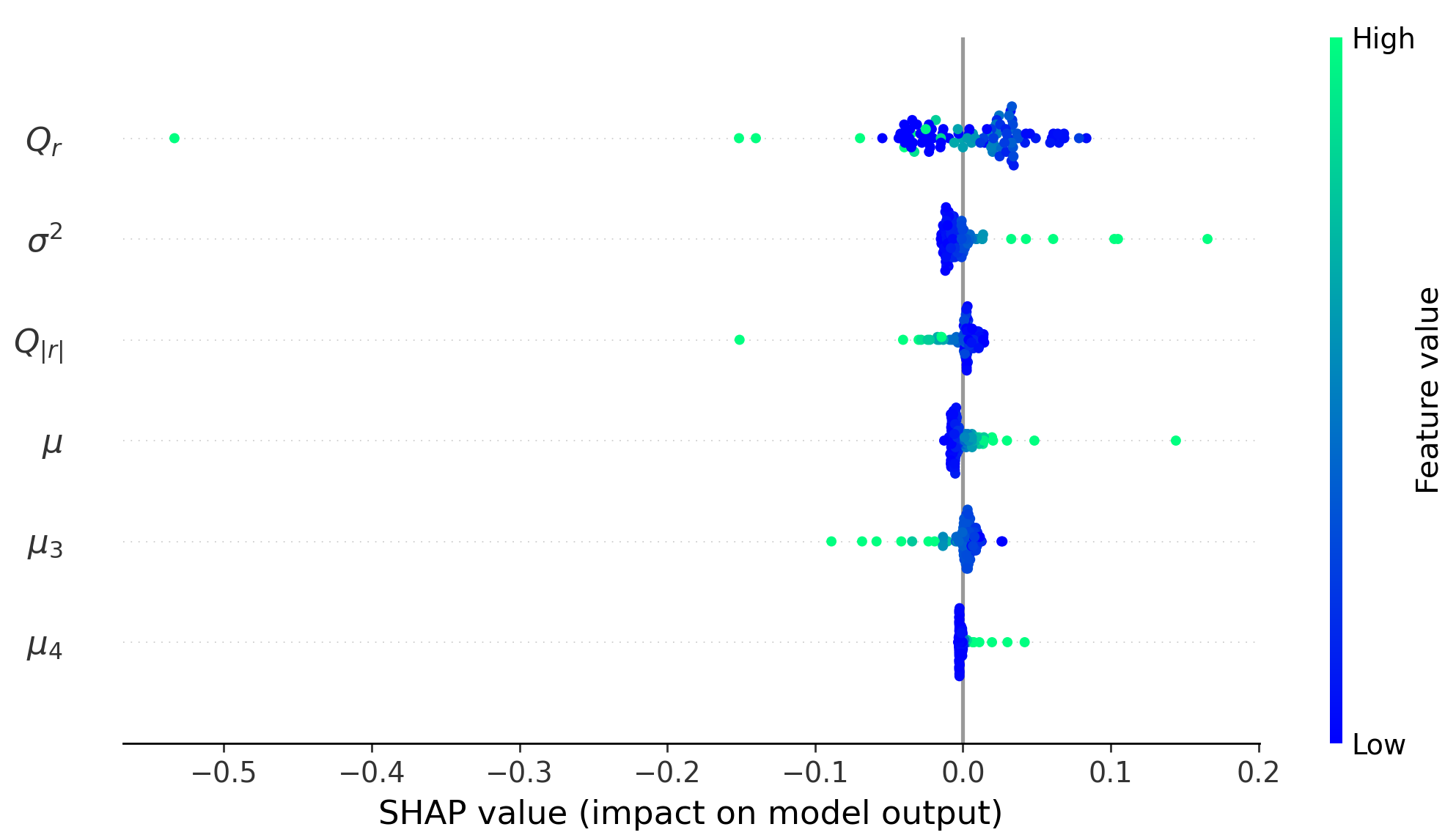}
    \caption{SHAP values for $s_p$}
    \end{subfigure}
    \begin{subfigure}[b]{0.49\textwidth}
    \includegraphics[width=\textwidth]{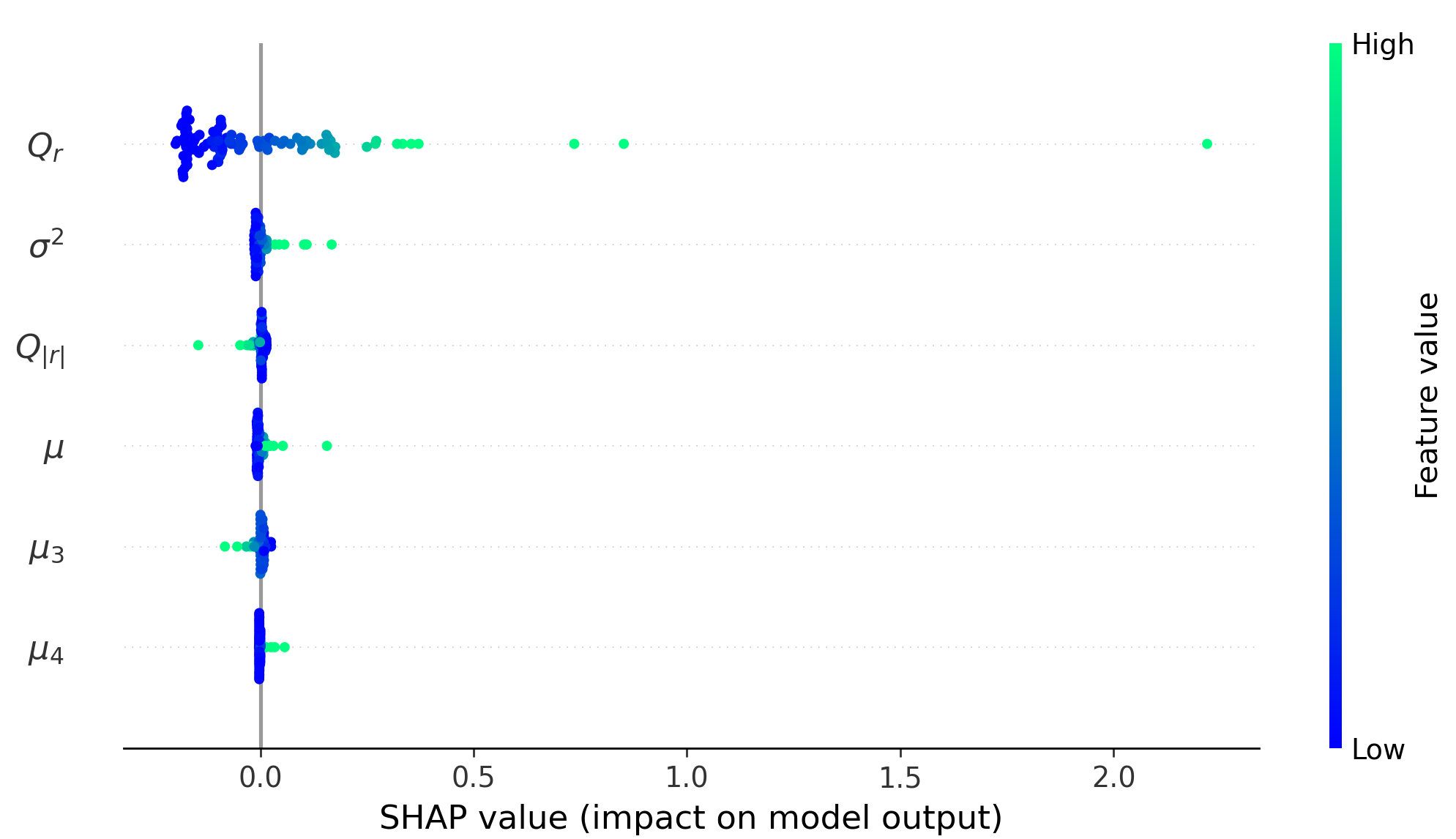}
    \caption{SHAP values for $s_d$}
    \end{subfigure}
    \begin{subfigure}[b]{0.49\textwidth}
    \includegraphics[width=\textwidth]{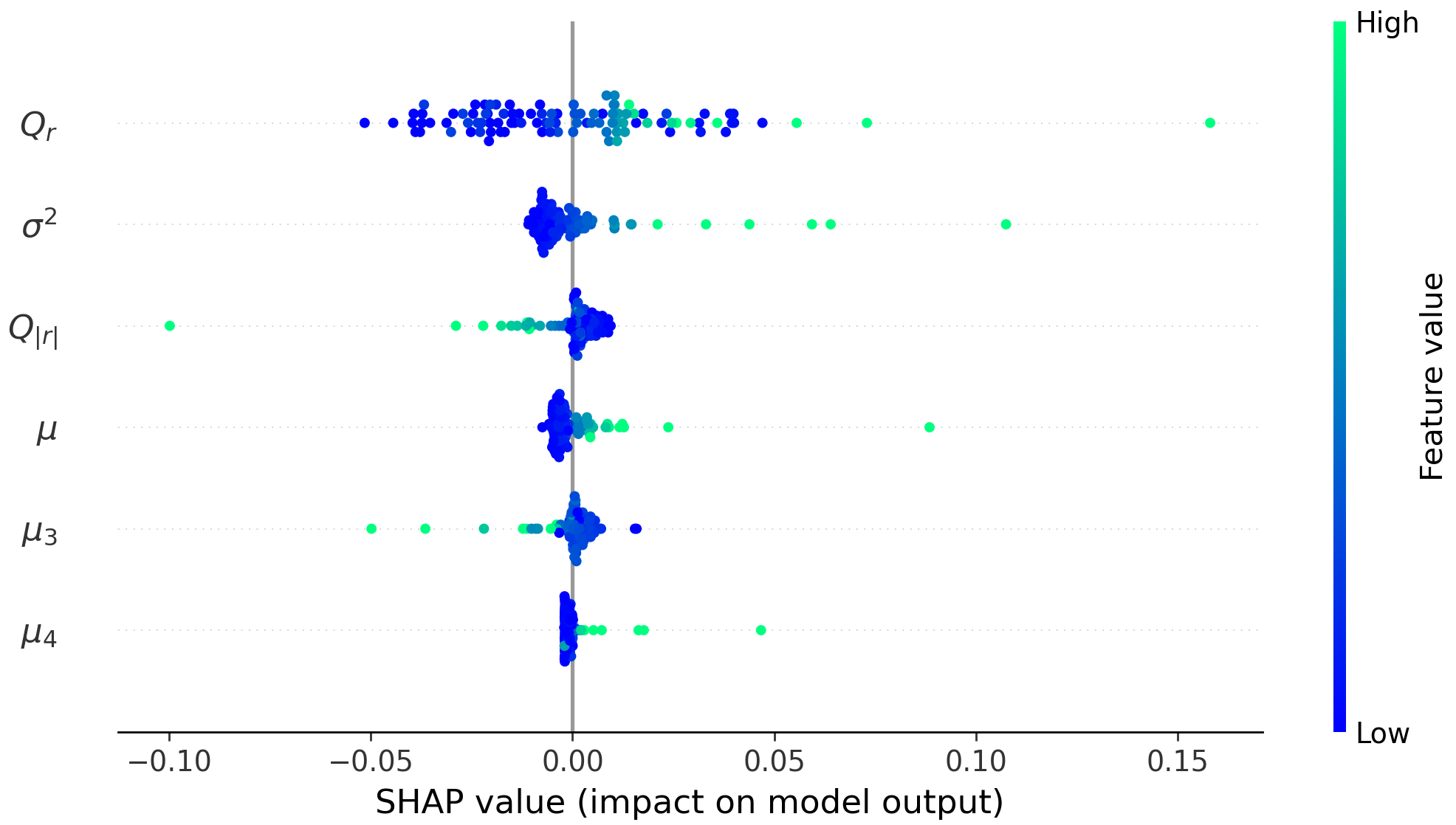}
    \caption{SHAP values for $s_a$}
    \end{subfigure}
    \begin{subfigure}[b]{0.49\textwidth}
    \includegraphics[width=\textwidth]{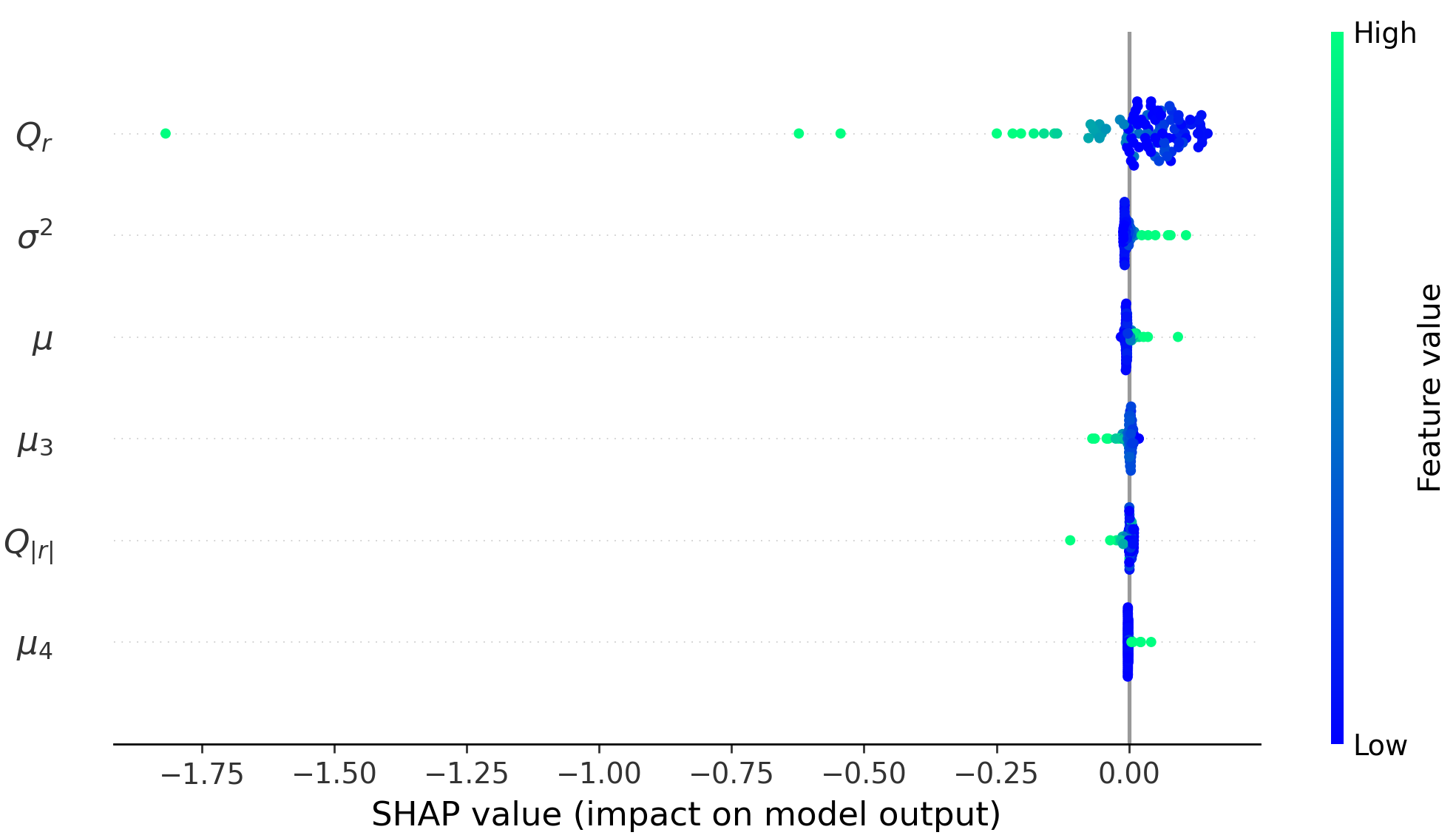}
    \caption{SHAP values for $\alpha^{\star}$}
    \end{subfigure}
    \label{shapley}
    \caption{Shapley Values for $s_p$, $s_d$, $s_a$ and $\alpha^{\star}$}
\end{figure}
In order to make a statement on the influence of features regarding the output, we regress the calculated statistics sequentially on $s_p$, $s_d$, $s_a$ and $\alpha^{\star}$. We train a neural network regression model with $n_{hl}=2$ hidden layers of $n_{ne}=10$ neurons each and use early stopping as regularization.
Figure 5 shows the Shapley values for the different characteristics of the data set where each represented by a dot. The corresponding plots sort the features regarding their importance from top to bottom and plot the Shapely value of each instance against its feature value indicated by the coloring from blue (low) to cyan (high). In Figure 5(a) we can see that the high first order autocorrelation of returns, denoted by $Q_r$, leads to a smaller $s_p$. Please remind at this point, that the smaller $s_p$, the better the generated data. This appears reasonable, since highly correlated time series are easier to predict than time series with little autocorrelation. 
For $Q_r$ multiple dots with high feature values (cyan) are visible. For the variance we see an opposite behavior, of high variance leading to a higher and thus worse predictive score. In addition, datasets with higher $\mu_4$ and $\mu$ result in a higher $s_p$, while $\mu_3$ has the tendency to improve $s_p$. What we can see here ist that the features mainly have either small positive or small negative effect and contribute largely only in case of extreme feature values.
For the discriminative score in Figure 5(b), we observe the same order in the feature importance. Interestingly, only the results of $Q_r$ reverse their effects, since one would expect the opposite effects consistently for all variables. But as mentioned above, the discriminative score measures a different aspect of the generation process. For the discriminative score, a higher variance leads to a higher discriminative score, which is reasonable, since a data set with $\sigma^2=0$ would be easily distinguishable from generated data, which naturally comes with some variance.\\
The Shapley value for the augmentation benefit $s_a$ are of particular interest for this paper. For $Q_r$ we can observe multiple interesting effects. the plot proposes that an average feature value increases the augmentation score. On the one hand, instances with a medium low level of $Q_r$ seem to improve the score. On the other hand elements with very low and very high feature values both increase the benefit of augmentation. Regarding $s_a$, this behavior is not observed for any other feature.
Strikingly, compared to $Q_r$, opposite effects can be observer for $Q_{\lvert r \rvert}$. It has the opposing effect not only in Figure 5(c) but also in the previous ones. Return distributions with high $\mu_3$ seem to harm $s_a$ while excess $\mu_4$ improves the augmentation benefit. However, the second and third moment are of relatively small importance compared to $Q_r$ and $\sigma$, unless they have high feature values.\\
Regarding the choice of $\alpha^{\star}$, we see in Figure 5(d) that smaller values of $Q_r$ have a small positive influence, but the higher values the more they reduce $\alpha^{\star}$. This means for data sets with high autocrrelation there exists the tendency to reduce the optimal level of $\alpha^{\star}$. The opposite applies for $\sigma^2$ and $\mu$ which have the tendency to favor higher fractions of generated data for higher features. It is clearly observable that all features except $Q_r$ have little effect on $\alpha^{\star}$ unless they take high values.
Throughout the evaluation of $s_p$, $s_d$, $s_a$, $\alpha^{\star}$ we see that the test statistic of the Ljung Box test provides most information regarding the Shapley values. The other features become only relevant once their values are high. However, we cannot make a deterministic statement which level of $\alpha$ is the best, but we can surely provide some guidance on whether an augmentation is likely to be successful and limit the search space for an exhaustive search of $\alpha$.
\newpage
\section{Conclusion}
This paper evaluates the image-based financial time series augmentation method for the first time. The new extended IRP encodes the initial time series as well as its intertemporal dynamics in a single image. In a sequence of tests the data augmentation increases the performance of return forecasting models substantially. This is valid for all considered data frequencies (daily, weekly, monthly, quarterly and yearly) where the model decreased the average forecasting error by 7\%. In a series of tests and multiple repetitions we observe an optimal amount of augmented data ranging from 17-25\% and the forecast error was reduced by 79\% of the datasets. Further, Shapley values identify important features such as low variance and high autocorrelation to have positive effects on the quality of the simulated returns indicated by the performance metrics. In addition, the distribution of optimal augmentation levels proposes a fraction of 19\% of synthetic data over all data frequencies. In a more far reaching study, the number of predictors in the analysis could be extended to account for drifts, seasonality and jump frequency. Further, a cross model comparison with other time series augmentation methods could provide useful benchmarks regarding the efficiency of the WGAN-GP trained on XIRP.
\newpage
\section{Appendix}
\begin{landscape}
\includepdf[pages=-,pagecommand={},angle=90,scale=.95]{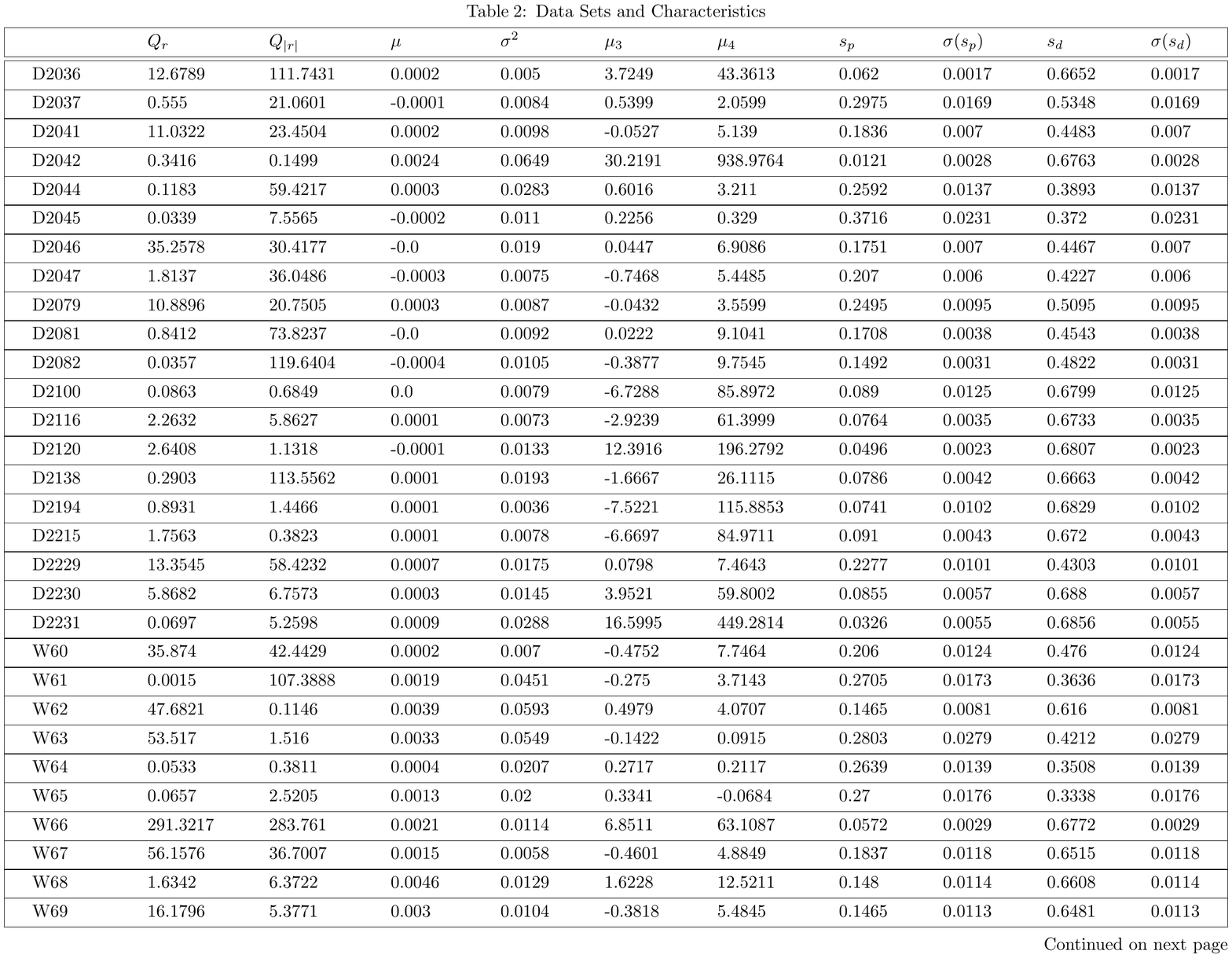}
\includepdf[pages=-,pagecommand={},angle=90,scale=.95]{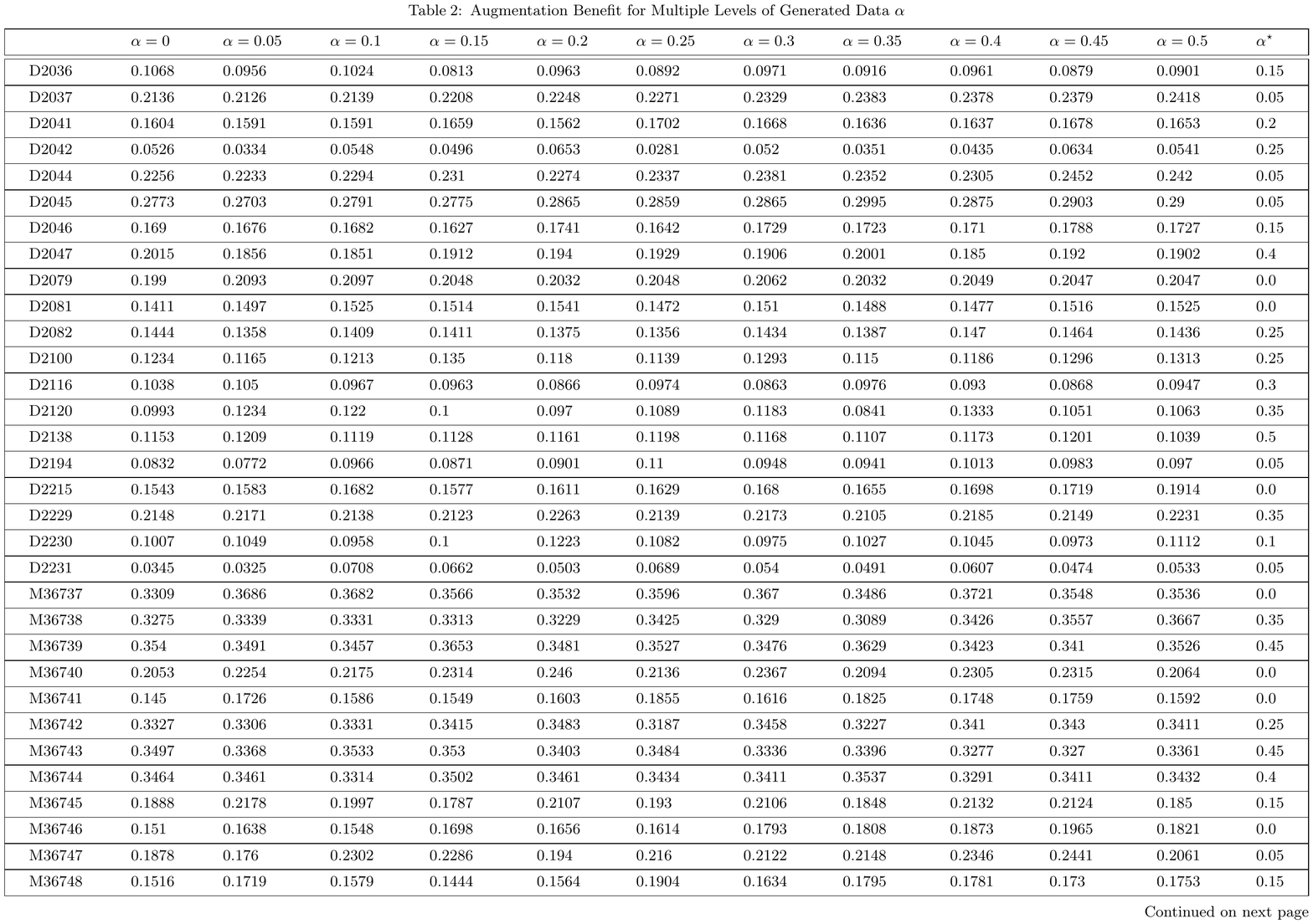}
\end{landscape}
\newpage

\bibliographystyle{plainnat}
\bibliography{references}
\end{document}